\documentclass[lettersize,journal]{IEEEtran}
\usepackage{amsmath,amsfonts}
\usepackage{amssymb}
\usepackage{algorithmic}
\usepackage{algorithm}
\usepackage{array}
\usepackage[caption=false,font=normalsize,labelfont=sf,textfont=sf]{subfig}
\usepackage{textcomp}
\usepackage{stfloats}
\usepackage{url}
\usepackage{verbatim}
\usepackage{graphicx}
\usepackage{cite}
\usepackage{tabularx}
\usepackage[pagebackref, breaklinks, colorlinks]{hyperref}
\usepackage{booktabs}
\usepackage{makecell}
\usepackage{bm}
\begin{document}

\title{PhysDrift: Bridging the Embodiment Gap\\in Humanoid Co-Speech Motion Generation}

\author{
	Zhangzhao Liang\textsuperscript{1}, Xiaofen Xing\IEEEauthorrefmark{1}\textsuperscript{1}, Mingyue Yang\textsuperscript{2}, Wenlve Zhou\textsuperscript{3}, Xiangmin Xu\IEEEauthorrefmark{1}\textsuperscript{3}\\
\textsuperscript{1}South China University of Technology \textsuperscript{2}DexForce Technology  \textsuperscript{3}Foshan University\\ \IEEEauthorrefmark{1}Corresponding Author
} 


%
\maketitle

\begin{figure*}[!t]
	\includegraphics[width=7.2in]{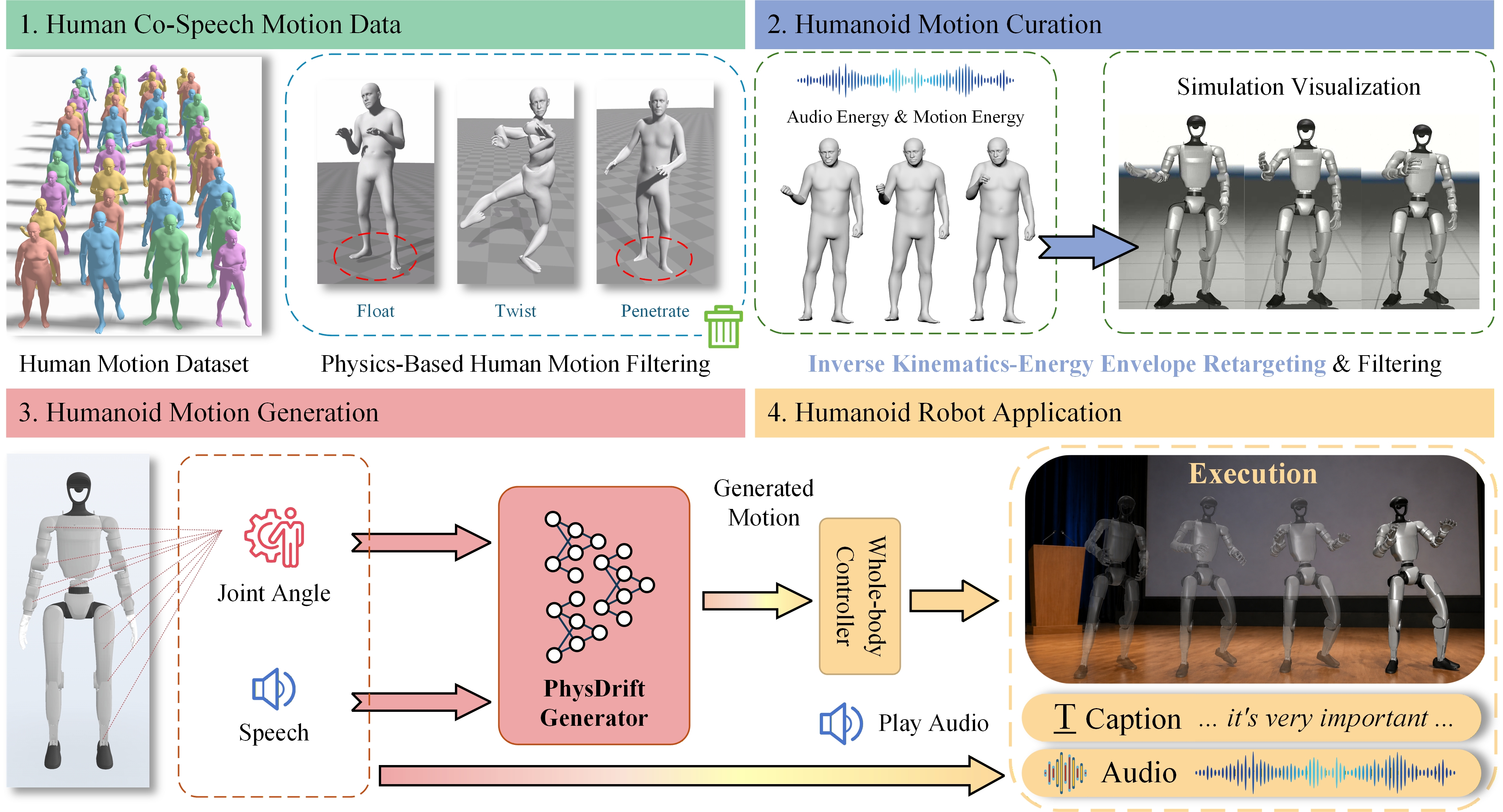}
	\caption{From human motion capture data to humanoid robot co-speech motion. The entire pipeline consists of four steps. First, data that clearly violates physical laws or severely distorts limbs in human motions is filtered out. Next, the proposed IK-EER maps human motions onto humanoid robots to obtain co-speech motions in the robot's native motion space, while further removing data that does not match the robot's attributes. Then, a PhysDrift generation model is trained using the robot’s co-speech motions. Finally, the whole-body controller takes the motion generated by PhysDrift as a reference to execute actions on the real robot.}
	\label{fig1}
\end{figure*}

\begin{abstract}
Humanoid robots require co-speech motions that are not only expressive and speech-aligned, but also physically executable under embodiment constraints. Existing co-speech generation pipelines are predominantly human-centric: motions are first generated in human-body representations such as SMPL-X and subsequently retargeted to humanoid robots. In this work, we identify a fundamental embodiment gap in this paradigm, where the mismatch between human motion manifolds and humanoid embodiment constraints disrupts embodiment consistency during motion transfer and physical execution. Through extensive analysis, we show that although retargeting can preserve coarse motion semantics, it significantly compresses motion diversity and weakens prosody-motion synchronization, limiting expressive humanoid behaviors. To address this problem, we first propose IK-EER, a prosody-preserving humanoid motion curation framework that jointly optimizes kinematic feasibility and speech-motion temporal alignment during retargeting. Building upon the curated robot-native motion dataset, we further introduce PhysDrift, an embodiment-aware co-speech motion generation framework that directly predicts executable humanoid joint trajectories from speech without relying on intermediate human-body representations. Unlike conventional human-centric pipelines, PhysDrift maintains embodiment consistency throughout both training and inference while incorporating physical regularization to stabilize robot motion dynamics. Extensive experiments and real-world humanoid deployment demonstrate that embodiment-aware robot-native generation substantially improves speech-motion alignment, physical plausibility, motion smoothness, inference efficiency, and real-time interaction capability. The results further reveal that robot-native motion representations are fundamentally more suitable than human-centric intermediates for embodied co-speech interaction in humanoid robots.
\end{abstract}

\begin{IEEEkeywords}
Co-Speech Motion, Retargeting, Motion Generation, Human-Robot Interaction.
\end{IEEEkeywords}

\section{Introduction}
\IEEEPARstart{H}{umanoid} robots are expected to engage in natural face-to-face interaction with humans \cite{I-2,I-3,I-4,I-5,I-6}, where speech is tightly coupled with expressive body motion. Co-speech motions play a critical role in this process by conveying emphasis, rhythm, emotion, and conversational intent beyond verbal content alone. Recent advances in generative modeling have substantially improved the realism and diversity of human co-speech motion synthesis \cite{livegesture,CoordSpeaker}. These approaches demonstrate impressive capability in learning speech-motion correspondence from large-scale human motion datasets and have shown promising results for virtual avatars and digital humans.

Despite this progress, transferring co-speech motion generation from virtual humans to physical humanoid robots remains fundamentally challenging. Similar to the motion of general humanoid robots, existing pipelines \cite{UniTracker,kungfubot} are predominantly human-centric: motion is first generated in human-body representations such as SMPL-X \cite{I-1} and subsequently retargeted onto humanoid embodiments through inverse kinematics or optimization-based motion transfer. While effective for animation, this paradigm implicitly assumes that human motion representations are compatible with humanoid embodiment constraints. However, humanoid robots differ substantially from humans in kinematic structure, joint limits, actuation capability, balance constraints, and feasible motion manifolds. As a result, motion representations learned in human-centric latent spaces are not naturally aligned with the physically executable motion space of humanoid robots \cite{I-12}.

In this work, we identify and formalize this discrepancy as an \textit{embodiment gap} in humanoid co-speech motion generation. Unlike prior works that primarily focus on motion retargeting accuracy or kinematic feasibility, we show that the embodiment gap manifests more fundamentally as a distribution mismatch between human motion manifolds and robot-executable motion manifolds. Our experiments reveal that although modern retargeting methods can largely preserve coarse motion semantics and avoid severe joint violations, the retargeting process significantly compresses motion diversity and weakens prosody-motion synchronization. Consequently, expressive speech-driven motions learned in human-centric representations become progressively distorted during humanoid reconstruction and physical execution.

A natural solution is to abandon human-centric intermediate representations and directly model humanoid co-speech behavior in robot joint space. However, robot-native generation introduces a new challenge. Without embodiment-aware constraints, highly expressive generative models, particularly flow-based models \cite{I-11}, can easily produce physically unstable motions with excessive jerk, contact artifacts, and joint-limit violations, despite achieving strong distributional metrics. Therefore, effective humanoid co-speech motion generation requires not only expressive generative capability, but also embodiment-consistent physical regularization.

To address these challenges, we propose PhysDrift, an embodiment-aware robot-native framework for humanoid co-speech motion generation. Instead of relying on intermediate human-body representations, PhysDrift directly predicts executable humanoid joint trajectories from speech. To construct high-quality robot-native training data, we further introduce Inverse Kinematics-Energy Envelope Retargeting (IK-EER), a prosody-preserving humanoid motion curation framework that jointly optimizes kinematic feasibility and speech-motion temporal alignment during retargeting. Building upon the curated dataset, PhysDrift incorporates embodiment-aware regularization to stabilize generated motion dynamics while preserving speech-motion alignment and real-time generation capability. Fig. \ref{fig1} provides an overview of the research presented in this paper.

Extensive experiments across motion quality, physical feasibility, speech alignment, and deployment efficiency demonstrate that embodiment-aware robot-native generation substantially outperforms conventional human-centric pipelines. In particular, our method achieves superior speech-motion synchronization, smoother physical dynamics, significantly faster inference speed, and more stable humanoid execution while maintaining expressive motion diversity. We further validate the proposed framework through real-world humanoid deployment, demonstrating robust real-time co-speech interaction capability.

Our contributions are summarized as follows:

\begin{itemize}
	\item We identify and formalize the embodiment gap in humanoid co-speech motion generation, showing that human-centric motion representations fundamentally disrupt embodiment consistency during retargeting and physical execution.
	
	\item We propose IK-EER, a prosody-preserving humanoid motion curation framework that jointly considers kinematic feasibility and speech-motion temporal alignment during retargeting.

	\item We propose PhysDrift, an embodiment-aware robot-native co-speech motion generation framework that directly predicts executable humanoid joint trajectories from speech without relying on intermediate human-body representations.
	
	\item We demonstrate through extensive experiments and humanoid deployment that embodiment-aware generation substantially improves physical plausibility, motion smoothness, speech alignment, inference efficiency, and real-time interaction capability for humanoid co-speech motion.
	
\end{itemize}

\section{Related Work}
In this section, we review prior work from three perspectives closely related to humanoid co-speech motion generation: human-centric co-speech generation, motion retargeting for humanoid robotics, and robot-native humanoid motion generation. Unlike conventional categorizations, we particularly focus on how existing methods model the relationship between motion representation and physical embodiment, which forms the core challenge addressed in this work.
﻿
\subsection{Human-Centric Co-Speech Motion Generation}
Recent progress in co-speech motion generation has been primarily driven by the computer vision and graphics communities. Existing approaches typically formulate the task as generating human body motion conditioned on speech signals, where motions are represented in human skeletal spaces or parametric body models such as SMPL-X.
Large-scale conversational motion datasets have substantially accelerated this research direction. BEAT \cite{II-I-1} and BEAT2 \cite{II-I-2} provide multilingual conversational motion capture datasets with synchronized speech and body motion annotations. ZeroEGGS \cite{II-I-4} further contributes high-fidelity expressive speaking styles for gesture synthesis. In parallel, large-scale Internet video datasets such as AVSpeech \cite{II-I-5} and the TED Gesture Dataset \cite{II-I-6} enable scalable learning of speech-driven gestures from unconstrained audio-visual data.
Building upon these datasets, recent generative models have achieved remarkable progress in producing realistic and semantically meaningful human gestures. HOP \cite{I-7} models multimodal interactions among speech, text, and gesture dynamics. SemGes \cite{I-8} improves semantic consistency through local-global constraints. DIDiffGes \cite{I-9} introduces an efficient decoupled diffusion framework for gesture generation, while MotionCraft \cite{I-10} employs unified diffusion transformers for multimodal whole-body motion synthesis.
Despite their impressive visual quality, these methods are fundamentally designed for digital humans rather than physically embodied humanoid robots. More importantly, their learned motion representations are intrinsically human-centric, assuming motion manifolds defined by human kinematics and morphology. Such representations do not explicitly account for embodiment-specific constraints including robot joint structures, actuator limitations, balance constraints, or physically executable humanoid motion spaces. Consequently, transferring these generated motions onto humanoid robots inevitably requires an additional embodiment transfer stage, introducing a mismatch between human motion representations and robot-executable behaviors.
﻿
\subsection{Motion Retargeting for Humanoid Robotics}
Motion retargeting aims to transfer human motion onto humanoid embodiments while preserving motion semantics and physical feasibility. In humanoid robotics, retargeting has become a practical strategy for leveraging large-scale human motion priors to improve robot motion generation and control.
Traditional humanoid motion generation approaches \cite{II-II-1,II-II-2,II-II-3,II-II-4,II-II-5} mainly rely on trajectory optimization or manually designed controllers. Although these methods achieve stable motion control, they often struggle to reproduce expressive and socially meaningful human motion dynamics. To address this limitation, recent retargeting frameworks attempt to transfer natural human movements onto humanoid embodiments.
Recent studies have significantly improved the physical plausibility of retargeted humanoid motion. Jeong et al. \cite{II-II-6} proposed a standardized retargeting framework addressing self-collision and contact consistency. Exbody \cite{II-II-7} introduced local joint mapping strategies for transferring human motion onto the Unitree H1 humanoid platform. Lu et al. \cite{II-II-8} refined inverse-kinematics-based upper-body retargeting to preserve natural motion characteristics under embodiment differences. Mao et al. \cite{II-II-9} further demonstrated scalable retargeting pipelines for converting Internet-scale human motions into executable humanoid datasets.
However, existing retargeting methods primarily optimize spatial pose reconstruction and physical feasibility while paying limited attention to embodiment consistency in speech-driven interaction. In particular, co-speech motions depends not only on pose accuracy but also on subtle temporal coupling between speech prosody and motion dynamics. Retargeting processes based on frame-wise inverse kinematics, smoothing, or projection into feasible robot subspaces often distort expressive motion distributions and weaken prosody-motion synchronization. As a result, motions that remain physically executable after retargeting may still lose expressive diversity and conversational naturalness during humanoid interaction.
﻿
\subsection{Robot-Native Humanoid Motion Generation}
More recent research has begun exploring robot-native humanoid motion generation frameworks that operate directly within robot action spaces instead of relying entirely on human-body intermediates. These methods aim to incorporate embodiment constraints during generation and control more explicitly.
Several approaches explore language-conditioned humanoid generation and interaction. Harmon \cite{II-III-1} combines human motion priors with vision-language reasoning for whole-body humanoid generation. Xu et al. \cite{II-III-2} proposed a text-driven humanoid motion generation framework on the NAO platform using angle-space representations and reinforcement learning controllers. Bao et al. \cite{II-III-3} introduced a hierarchical framework integrating intention reasoning and diffusion-based social gesture generation for humanoid interaction.
Parallel advances have also emerged in imitation learning and diffusion-policy-based humanoid control. Ze et al. \cite{II-III-4} combined teleoperation and diffusion policies for full-body humanoid skill learning. HOVER \cite{II-III-5} proposed a neural whole-body humanoid controller, while ManiDP \cite{II-III-6} introduced manipulability-aware diffusion policies for bimanual humanoid manipulation.
Although these methods operate more closely to robot embodiments, most focus on locomotion, manipulation, or general task-oriented motion rather than speech-driven social interaction. Moreover, many approaches still rely on human-derived motion priors, reference trajectories, or intermediate representations during training. Consequently, the problem of jointly preserving speech alignment, expressive motion dynamics, and embodiment-consistent physical execution remains insufficiently explored in humanoid co-speech motion generation.
﻿
\subsection{Summary}
Existing co-speech motion generation methods predominantly rely on human-centric motion representations and post-hoc retargeting pipelines, while existing humanoid motion generation approaches mainly focus on task-oriented behaviors rather than expressive speech-driven interaction. As a result, current methods lack an explicit mechanism for maintaining embodiment consistency between speech prosody, expressive motion dynamics, and physically executable humanoid behavior. In contrast, this work proposes an embodiment-aware robot-native co-speech generation framework that jointly considers humanoid motion curation, embodiment-consistent motion representation, and physically grounded speech-driven generation directly within humanoid joint space.

\begin{figure*}[!t]
	\includegraphics[width=7in]{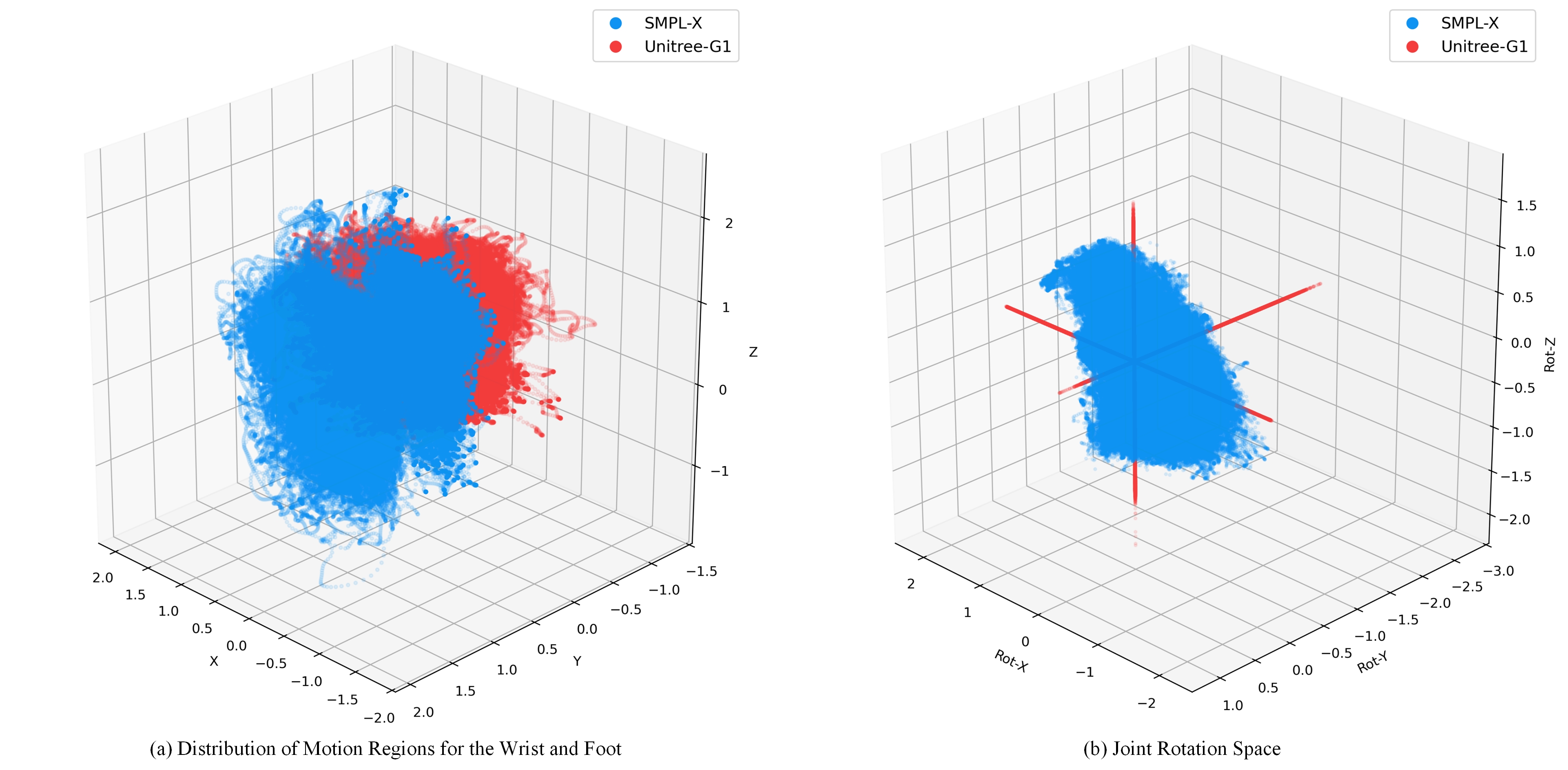}
	\caption{The feasible region of the end effectors and the joint rotation space between SMPL-X and Unitree-G1 are illustrated. In (a), the end effectors are the wrist and foot, which are common to both G1 and SMPL-X. (b) shows the joint rotation space of the shoulders, wrists, and hips for SMPL-X and Unitree-G1.}
	\label{fig:eef gap}
\end{figure*}

\section{Embodiment Gap Analysis and Problem Formulation}
\textbf{Embodiment Gap.} The primary distinction between SMPL-X pose representation and robotic joint configuration lies in the nature of rotational freedom and physical constraints. For a given joint, SMPL-X uses an axis-angle representation:
\begin{equation}
	\mathbf{r} = \theta \mathbf{\hat{u}},
\end{equation}
where $\mathbf{\hat{u}}\in \mathbb{R}^3$ denotes the unit rotation axis satisfying $\|\mathbf{\hat{u}}\|=1$, $\|\cdot\|$ denotes the Euclidean norm, and $\theta$ is the rotation magnitude. This formulation allows arbitrary 3D orientations without consideration of mechanical limits or collisions, effectively sampling from the mathematical rotation group $SO(3)$. Consequently, SMPL-X can represent poses that are physically impossible for a real human or a robot, including extreme rotations or self-intersecting limbs. 

In contrast, a robot's joint configuration is defined as
\begin{equation}
	q_i \in [q_i^{\min}, q_i^{\max}], \quad i = 1, \dots, N_{\text{joints}},
\end{equation}
where each $q_i$ corresponds to a mechanically constrained degree of freedom (DoF), often restricted to rotation about a fixed axis. $q_i^{\min}$ and $q_i^{\max}$ denote the rotation range of the joint. These limits arise from actuator capabilities, structural design, collision avoidance, and stability requirements. Unlike SMPL-X, robotic joints cannot arbitrarily rotate, and multiple axis rotations require solving inverse kinematics under constraints.

The embodiment gap is summarized as Table \ref{tab:structure} highlighting that while SMPL-X represents idealized human-like orientations, robotic joints encode physically realizable actuator configurations.
\begin{table}
	\centering
	\caption{The difference in body structure between SMPL-X and Unitree-G1}
	\renewcommand{\arraystretch}{1.25}
	\label{tab:structure}
	\resizebox{\linewidth}{!}{
		\begin{tabular}{l|c|c|c|c}
			\toprule
			Body & \# Joints & DoF & Rotation Rep. & Rotation Restrict \\
			\midrule
			SMPL-X & 24 Body + 30 Fingers & 162 & Axis-Angle & $\times$ \\
			G1 & 1 Root + 29 Body & 29 & Joint Angle & \checkmark \\
			\bottomrule
	\end{tabular}}
\end{table}

\textbf{Problem Formulation.} Humanoid co-speech motion generation aims to synthesize expressive body motion conditioned on speech signals while maintaining physical executability on humanoid platforms \cite{II-I-2,II-III-1}. Given an input speech sequence $A=\{a_t\}_{t=1}^{T},$
the goal is to generate a corresponding humanoid motion sequence $Q=\{q_t\}_{t=1}^{T}$, where \(q_t \in \mathbb{R}^{D}\) denotes the humanoid joint configuration at time step \(t\), including whole-body joint rotations and root motion parameters.

Due to the lack of in-depth research on the co speech motion of native humanoid robots. Therefore, a more widespread approach is first generate motion in human-centric representations such as skeletal pose spaces or parametric body models, and subsequently transfer the generated motions onto humanoid embodiments through retargeting pipelines. This process can be summarized as
\begin{equation}
	A \xrightarrow{\text{Stage 1}} H \xrightarrow{\text{Stage 2}} Q,
\end{equation}
where \(H\) denotes motion represented in a human-body joint space and the second stage corresponds to humanoid retargeting.

Although this paradigm has demonstrated strong visual performance for digital humans, it implicitly assumes that human motion representations are compatible with humanoid embodiment constraints. However, humanoid robots differ fundamentally from humans in morphology, kinematic structure, joint limits, actuation capability, and balance dynamics \cite{II-II-7, humanmimic}. As a result, the feasible humanoid motion space does not coincide with the motion manifold learned from human motion data. Shown in Fig. \ref{fig:eef gap} are the feasible region of the end effectors and the joint rotation manifold between SMPL-X and Unitree-G1. Bridging this gap requires mapping unconstrained human poses to constrained robot motions, typically via inverse kinematics with joint limits and collision avoidance. 
﻿
We denote the human motion manifold as \(\mathcal{M}_h\) and the physically executable humanoid motion manifold as \(\mathcal{M}_r\). Existing human-centric generation methods effectively learn distributions over \(\mathcal{M}_h\), while humanoid execution is constrained within \(\mathcal{M}_r\). Since $\mathcal{M}_h \neq \mathcal{M}_r$, retargeting inevitably becomes a projection process from human motion space into the feasible humanoid motion subspace:
\begin{equation}
	\mathcal{R}: \mathcal{M}_h \rightarrow \mathcal{M}_r.
\end{equation}
This discrepancy forms what we define as the \textit{embodiment gap} in humanoid co-speech motion generation.

Importantly, the embodiment gap does not necessarily manifest as severe kinematic failure. Modern retargeting methods can often maintain basic physical feasibility and preserve coarse motion semantics. However, the projection from \(\mathcal{M}_h\) to \(\mathcal{M}_r\) inevitably compresses expressive motion distributions and alters fine-grained temporal dynamics. In co-speech interaction, where gesture rhythm, motion energy, and speech prosody are tightly coupled, such distortions accumulate into perceptually significant degradation of conversational naturalness. This phenomenon is particularly evident in expressive gestures. Human co-speech motions often rely on subtle upper-body coordination, asymmetric arm dynamics, and temporally localized motion emphasis synchronized with speech rhythm \cite{EmotionGesture,MambaGesture2}. During retargeting, these high-frequency expressive components are frequently smoothed or projected into more conservative feasible robot motions, reducing motion diversity and weakening prosody-motion synchronization even when the resulting motions remain physically executable.

Our experimental analysis further confirms this observation. While retargeted motions retain relatively similar semantic alignment scores, motion diversity decreases substantially after retargeting across multiple generation methods. These results suggest that the primary limitation of human-centric pipelines is not merely physical feasibility, but the loss of embodiment consistency between expressive speech-driven motion dynamics and humanoid execution constraints.

A natural solution is therefore to directly model humanoid co-speech motion within robot-native action spaces \cite{II-III-1}. Instead of learning human motion distributions followed by embodiment transfer, we aim to directly learn the conditional distribution $p(Q|A)$, where motion generation and embodiment constraints are jointly modeled within humanoid joint space. This formulation motivates the proposed PhysDrift framework, which combines embodiment-aware motion curation and robot-native speech-conditioned motion generation to preserve both expressive dynamics and physical executability.

\section{Method}

\subsection{Overview}

The embodiment mismatch manifests in two aspects. First, human-centric representations contain large regions that are unreachable for humanoid embodiments, leading to unstable or over-smoothed motions after retargeting. Second, frame-wise inverse kinematics and motion smoothing operations distort subtle temporal patterns synchronized with speech rhythm and emphasis, which are essential for natural co-speech behavior.

To address these issues, we reformulate humanoid co-speech motion generation as a direct robot-space generation problem. Instead of treating robotic embodiment as a post-processing constraint, embodiment consistency is explicitly modeled during both dataset construction and motion generation. Our framework consists of two tightly coupled components. First, we propose IK-EER, an embodiment-aware humanoid motion construction framework that converts human co-speech motions into robot-native supervision while preserving speech-motion synchronization and physical feasibility. Second, we introduce PhysDrift, a robot-native speech-driven motion generation framework that directly learns executable humanoid motion distributions in joint space without relying on intermediate human-body representations.

\subsection{Embodiment-Aware Motion Construction}

\begin{figure}[!t]
	\includegraphics[width=3.5in]{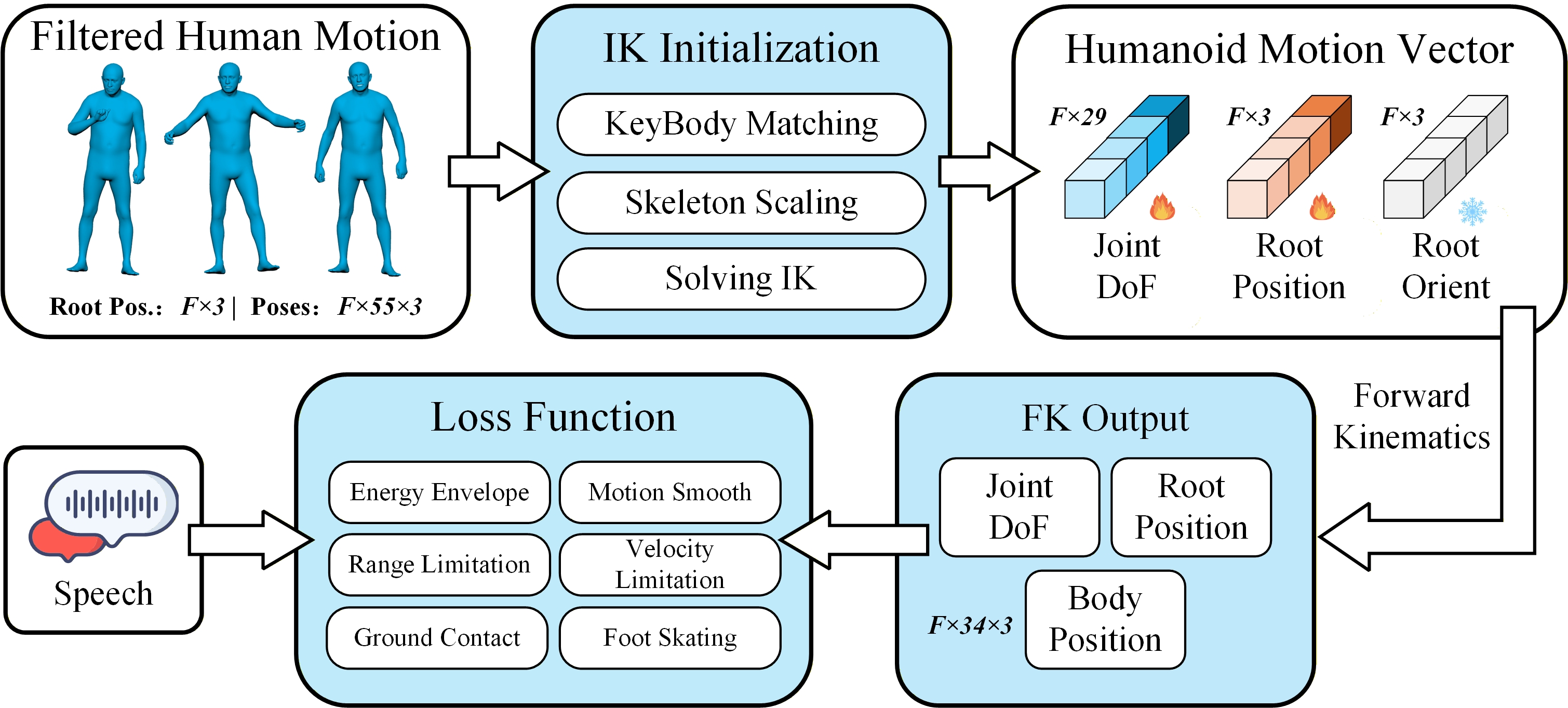}
	\caption{Pipeline of IK-EER. ``Joint DoF'' and ``Root Position'' are optimized through gradient-based motion refinement, while the root orientation is directly inherited from the human motion sequence.}
	\label{fig2}
\end{figure}

Constructing a high-quality humanoid co-speech motion dataset is critical for robot-native motion generation. However, due to the lack of available robot co-speech motion, a common method is to retarget from human motion data to the robot body. Existing retargeting pipelines \cite{III-II-1,III-II-2,PHC} primarily optimize spatial pose reconstruction fidelity, implicitly assuming that physically feasible reconstruction is sufficient for downstream learning. However, for co-speech motions, preserving temporal synchronization between speech prosody and motion dynamics is equally important. Motions that are physically executable but temporally inconsistent with speech often appear socially unnatural during interaction.

To address this issue, we propose IK-EER, an embodiment-aware motion construction framework designed not merely for pose transfer, but for generating robot-native supervision suitable for humanoid co-speech learning. The details of IK-EER is shown in Fig. \ref{fig2}.

Starting from the BEAT2 dataset, We removed physically infeasible sequences from the source data (such as floating, twisted limbs, ground penetration). Instead of directly adopting SMPL-X parameters as training targets, all motions are transformed into the native joint space of the humanoid robot. For each frame, robot joint configurations are initialized through sparse human-to-robot keypoint correspondences using inverse kinematics. Let $\mathbf{q}$ denote robot joint angles. The initialization objective is formulated as
\begin{equation}
	\min_{\mathbf{q}}
	\sum_i
	w_i^p
	\|
	\mathbf{p}_i^{target}
	-
	\mathbf{p}_i(\mathbf{q})
	\|^2
	+
	w_i^r
	\|
	\mathbf{R}_i^{target}
	-
	\mathbf{R}_i(\mathbf{q})
	\|_F^2 ,
\end{equation}
where $\mathbf{p}_i$ and $\mathbf{R}_i$ denote the position and orientation of the corresponding robot joint \textit{i}. $w_{i}$ is the weighting factor and \textit{target} denotes the target human motion position or orientation.

Although inverse kinematics provides feasible initialization, conventional retargeting pipelines often weaken the intrinsic coupling between speech rhythm and motion dynamics due to smoothing and kinematic corrections. Human co-speech motions naturally exhibit strong correlations between acoustic emphasis and motion intensity. However, this relationship is highly sensitive to temporal distortion during embodiment projection.

To preserve this synchronization, we introduce an Energy Envelope (EE) objective. The synchronization objective is defined using normalized cross-correlation (NCC) \cite{ncc} bewteen motion energy $E_m$ and audio energy $E_a$:
\begin{equation}
	\mathcal{L}_{EE}
	=
	1-
	\mathrm{NCC}(E_m,E_a).
\end{equation}

Due to the fact that movement speed is an important manifestation of motion rhythm, given a motion sequence $\mathbf{m}$, motion energy is estimated from joint velocities:

\begin{equation}
	E_m(t)
	=
	\left\|
	\frac{\mathbf{m}_{t+1}-\mathbf{m}_{t-1}}
	{2\Delta t}
	\right\|_2 .
\end{equation}

Similary, the raw waveform be denoted as $\mathbf{a} \in \mathbb{R}^{B \times L}$. We first compute an $80$-bin log-Mel spectrogram using an short-time Fourier transform, yielding $\mathbf{S} \in \mathbb{R}^{B \times 80 \times T}$. The frame-level log-Mel energy is then defined as

\begin{equation}
	E_{a}(t)
	= 
	\log\left( \sum_{m=1}^{80} \mathbf{S}_{:,m,t} + \epsilon \right),
\end{equation}
where $\epsilon = 10^{-6}$. Both signals are normalized before alignment. 

In addition, physical regularization terms $\mathcal{L}_{phys}$ \cite{phuma} are incorporated to enforce embodiment feasibility, including joint-limit penalties, foot-contact consistency, and skating suppression:

\begin{equation}
	\mathcal{L}_{motion}
	=
	\mathcal{L}_{EE}
	+
	\mathcal{L}_{phys}.
\end{equation}

Unlike conventional retargeting methods that prioritize geometric pose reconstruction, IK-EER explicitly prioritizes embodiment-consistent temporal dynamics. More importantly, IK-EER serves as a humanoid co-speech motion \emph{curation framework} rather than merely a retargeting algorithm. Since robot-native generation requires large-scale physically executable humanoid motion supervision, preserving speech-motion synchronization during humanoid reconstruction becomes essential for downstream learning. The resulting dataset provides robot-native motion distributions that simultaneously preserve embodiment feasibility and conversational expressiveness.

\subsection{Robot-Native Motion Representation}
Existing co-speech motion generation methods typically operate on human-centric representations such as SMPL-X parameters or discretized latent motion tokens. Although these representations are suitable for digital human animation, they introduce substantial embodiment mismatch for humanoid robots.

From the perspective of motion manifolds, human-centric joint spaces contain large regions that are unreachable for robot embodiments due to differences in kinematic topology, actuation structure, and joint constraints. Consequently, learning in human representation space inevitably introduces representation-level embodiment inconsistency.

To eliminate this mismatch, we directly represent motions in the native control space of the humanoid robot. For each frame, the motion state is defined as

\begin{equation}
	\mathbf{x}^{(t)}
	=
	[
	\mathbf{r}_{root}^{(t)},
	\theta_1^{(t)},
	\dots,
	\theta_N^{(t)},
	\mathbf{p}_{root}^{(t)}
	],
	\label{config_space}
\end{equation}
where $\mathbf{r}_{root}$ denotes the root orientation represented using continuous 6D rotation \cite{6d}, $\theta_i$ are robot joint angles, $\mathbf{p}_{root}$ represents the global root position, and superscript \textit{(t)} denotes the timestep \textit{t}.

This formulation establishes a one-to-one correspondence between generated motion trajectories and executable motor commands, removing the need for intermediate human-body reconstruction or post-hoc retargeting. More importantly, robot-native representation constrains motion learning directly within the feasible humanoid motion manifold, substantially reducing embodiment inconsistency during generation.

\subsection{PhysDrift}
\begin{figure*}[!t]
	\centering
	\includegraphics[width=7in]{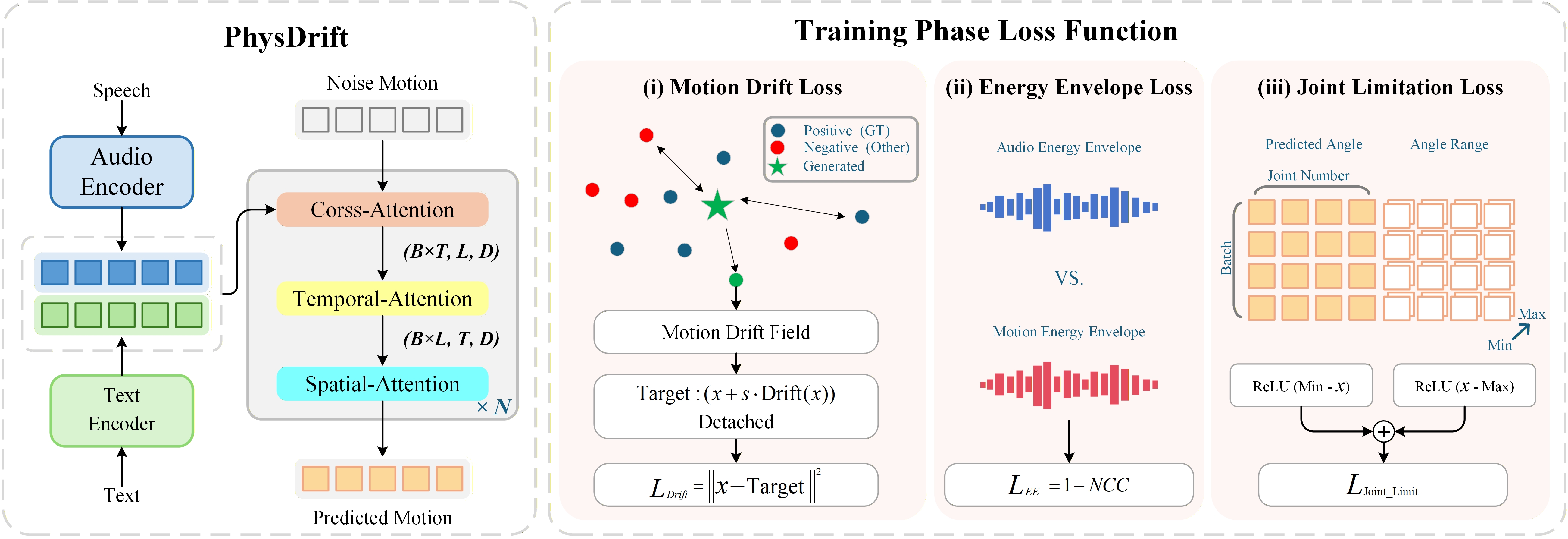}
	\caption{The proposed PhysDrift pipline. Left: the model architecture. Right: the loss function used during the training phase. Speech and text will be separately represented by modal encoders, and then fused with noisy motion sequences through cross-attention. Finally, the model outputs the corresponding predicted motion sequences.}
	\label{physdirft}
\end{figure*}
Building upon the robot-native motion representation, we propose PhysDrift, a direct speech-to-motion generation framework that operates entirely in humanoid joint space. Unlike conventional co-speech generation pipelines that first model human motions and subsequently adapt them to robotic embodiments, PhysDrift directly learns the distribution of executable humanoid motions, thereby eliminating the embodiment gap introduced by intermediate human-body representations. Fig. \ref{physdirft} shows the pipeline of PhysDrift.

Recent advances in generative modeling have demonstrated the effectiveness of diffusion models and Flow Matching for motion synthesis. In particular, Flow Matching learns a continuous transport process by estimating a velocity field
\begin{equation}
	\frac{d\mathbf{x}_t}{dt}
	=
	v_\theta(\mathbf{x}_t,t),
	\label{eq:flow_ode}
\end{equation}
which progressively transforms samples from a simple prior distribution into samples from the target motion distribution \cite{fm}. Compared with diffusion models, Flow Matching (FM) significantly reduces inference cost because the generation process can be formulated as deterministic trajectory transport.

However, we identify that directly learning velocity fields introduces a previously overlooked form of embodiment gap when applied to humanoid co-speech motion generation. Velocity-field learning implicitly assumes that generated trajectories will be decoded or smoothed before execution. This assumption is generally valid for digital human animation, where latent motions are reconstructed through VQVAE \cite{vqvae} Decoder or rendering pipelines that naturally suppress high-frequency artifacts. In contrast, humanoid robots directly execute generated joint trajectories through motor controllers. Consequently, small velocity estimation errors are accumulated through temporal integration:
\begin{equation}
	\mathbf{x}_{t+\Delta t}
	=
	\mathbf{x}_t
	+
	\int_t^{t+\Delta t}
	v_\theta(\mathbf{x},\tau)d\tau,
\end{equation}
causing local oscillations to propagate into large acceleration fluctuations. As a result, velocity-based transport often produces excessive motioaan jerk, unstable joint accelerations, and physically undesirable behaviors. This issue becomes particularly severe for co-speech motions, where subtle rhythmic variations synchronized with speech prosody create high-frequency motion components that are easily amplified during velocity integration.

From an embodiment perspective, the root cause is that velocity fields are defined in the tangent space (the vector space formed by all possible ``directions" or ``velocities" at a certain point on a manifold) of the motion manifold
\begin{equation}
	v_\theta :
	\mathcal{X}
	\rightarrow
	T\mathcal{X},
\end{equation}
where $T\mathcal{X}$ denotes the tangent bundle of the motion manifold. Whereas humanoid execution is performed directly in configuration space (Eq. (\ref{config_space})). Therefore, minimizing velocity transport error does not necessarily minimize the execution error of robot motions since robot execution is ultimately performed in configuration space rather than velocity space. This mismatch constitutes a generation-level embodiment gap that remains even after robot-native motion representations are adopted.

We address this issue by reformulating motion generation as a manifold attraction problem rather than a velocity transport problem, which leads to the proposed Motion Drift Field (MDF). To better understand its relationship with FM, we revisit the transport dynamics learned by velocity-based generative models.

The continuous dynamics in Eq. (\ref{eq:flow_ode}) can be discretized as
\begin{equation}
	\mathbf{x}_{t+\Delta t}
	=
	\mathbf{x}_t
	+
	\Delta t \,
	v_\theta(\mathbf{x}_t,t).
\end{equation}
Therefore, FM can be interpreted as repeatedly predicting infinitesimal transport directions and accumulating them through numerical integration. While this formulation provides an accurate approximation of continuous transport trajectories, it requires the model to learn dynamics in tangent space rather than the robot execution space.

Instead of learning infinitesimal transport directions, PhysDrift directly models the accumulated displacement between a generated sample and the feasible humanoid motion manifold. Let
\begin{equation}
	\Delta \mathbf{x}
	=
	\mathbf{x}^{*}
	-
	\mathbf{x},
\end{equation}
where $\mathbf{x}^{*}$ denotes an equilibrium state on the target humanoid motion manifold. The proposed MDF directly estimates this displacement:
\begin{equation}
	\mathbf{D}(\mathbf{x})
	\approx
	\mathbf{x}^{*}
	-
	\mathbf{x}.
\end{equation}
From this perspective, MDF can be viewed as a one-step relaxation formulation of velocity transport. Rather than explicitly learning tangent-space dynamics and integrating them over time, PhysDrift directly learns the correction required to move a sample toward the target motion manifold. Consequently, the transport process is implicitly absorbed into the network parameters during training.

Let $p$ denote the distribution of robot-native motions constructed by IK-EER and let $q$ denote the generated motion distribution. For a motion sample $\mathbf{x}$, the MDF is formally defined as
\begin{equation}
	\mathbf{D}_{p,q}(\mathbf{x})
	=\dfrac{1}{Z_{p}Z_{q}} 
	\mathbb{E}_{p,q}
	\left[
	\tilde{k}(\mathbf{x},\mathbf{y}^{+})
	\tilde{k}(\mathbf{x},\mathbf{y}^{-})
	(\mathbf{y}^{+}-\mathbf{y}^{-})
	\right],
\end{equation}

\begin{equation}
	Z_{p}(\mathbf{x})
	=
	\mathbb{E}_{p}
	\left[
	\tilde{k}(\mathbf{x},\mathbf{y}^{+})
	\right]],
\end{equation}

\begin{equation}
	Z_{q}(\mathbf{x})
	=
	\mathbb{E}_{q}
	\left[
	\tilde{k}(\mathbf{x},\mathbf{y}^{-})
	\right],
\end{equation}

\begin{equation}
\tilde{k}(\mathbf{x}, \mathbf{y}) = \exp\left(-\frac{1}{\tau}\|\mathbf{x} - \mathbf{y}\|\right),
\end{equation}
where $\mathbf{y}^{+}\sim p$, $\mathbf{y}^{-}\sim q$. In order to better adapt to the robot's motion space, we have redefined  $\|\cdot\|$ as a set of structured distances to accommodate different motion representations in Eq. (\ref{config_space}), rather than $L_2$ distance used in \cite{drift}. Specifically, for the root orientation $\mathbf{r}_{root}$, we reconstruct the rotation using the 6D representation and map it to $SO(3)$. The distance between two orientations $d_{\text{rot}}$ is measured using the geodesic distance on $SO(3)$. 
\begin{equation}
	\label{eq:so3_dist}
	d_{\text{rot}}
	= \arccos \Bigg( \frac{ \mathrm{tr} \Big( (\mathbf{R}^{(1)})^\top \mathbf{R}^{(2)} \Big) - 1 }{2} \Bigg),
\end{equation}
where $\mathbf{R}^{(1)}, \mathbf{R}^{(2)} \in SO(3)$ are the corresponding root rotation matrices. $\mathrm{tr}(\cdot)$ represents the trace of the matrix. For joint angle $\theta_i$, we compute rotational differences using the periodic distance between joint angles. 
\begin{equation}
	\label{eq:joint_angle_dist}
	d_{\text{joint}}
	= \sqrt{ \sum_{i=1}^{n} \left( \min\left( \bigl| \theta_i^{(1)} - \theta_i^{(2)} \bigr|,\; 2\pi - \bigl| \theta_i^{(1)} - \theta_i^{(2)} \bigr| \right) \right)^2 }
\end{equation}
For the root position $\mathbf{p}_{root}$, we measure positional errors $d_{\text{pos}}$ using the $L_2$ distance. In summary, 
\begin{equation}
\|\mathbf{x} - \mathbf{y}\| = d_{\text{rot}} + d_{\text{joint}} +  d_{\text{pos}}.
\end{equation}

The corresponding training objective minimizes the residual magnitude of the drift field:
\begin{equation}
	\mathcal{L}_{drift}
	=
	\left\|
	\mathbf{x}
	-
	\mathrm{sg}
	\left(
	\mathbf{x}
	+
	\mathbf{D}_{p,q}(\mathbf{x})
	\right)
	\right\|_2^2,
\end{equation}
where $\mathrm{sg}(\cdot)$ denotes the stop-gradient operator. The proposed formulation naturally possesses several desirable theoretical properties.

\textbf{Proposition 1 (Equilibrium Property).}
When the generated distribution matches the target distribution, the expected drift vanishes:
\begin{equation}
	\mathbf{D}_{p,q}(\mathbf{x})
	=
	0.
\end{equation}
This indicates that the target humanoid motion manifold corresponds to a stationary equilibrium of the learning dynamics. Furthermore, the kernelized formulation induces a local attraction process. Since the similarity kernel suppresses long-range interactions and emphasizes nearby motion samples, the resulting dynamics behave similarly to a contraction mapping in local neighborhoods. Consequently, generated motions are progressively attracted toward feasible humanoid motion regions while avoiding abrupt trajectory jumps.

An important consequence of the proposed attraction dynamics is that motion corrections are predicted directly in the robot execution space. Consequently, PhysDrift avoids derivative amplification caused by high-frequency motion components, substantially reducing motion jerk and improving temporal smoothness. This property is particularly important for physically embodied co-speech motion, where motion stability strongly influences interaction quality.

To further preserve embodiment consistency during generation, we incorporate physical feasibility and speech-motion synchronization constraints. The final training objective is defined as
\begin{equation}
	\mathcal{L}
	=
	\mathcal{L}_{drift}
	+
	\mathcal{L}_{Joint\_limit}
	+
	\mathcal{L}_{EE},
\end{equation}
\begin{equation}
	\mathcal{L}_{Joint\_limit}
	=
	\mathrm{ReLU}(q^{min} - \mathbf{x}) + \mathrm{ReLU}(\mathbf{x} - q^{max}).
\end{equation}

\subsection{Inference}

The proposed formulation naturally enables one-step generation. Conventional diffusion models require iterative denoising, while FM requires numerical integration of learned velocity trajectories. In contrast, PhysDrift directly learns the equilibrium correction that moves a sample toward the target humanoid motion manifold.

Because the transport dynamics are absorbed into the learned MDF during training, the multi-step transport process is compressed into a single network evaluation. Therefore, inference does not require iterative denoising, trajectory rollout, or numerical ordinary differential equation integration.

Given speech features $\mathbf{A}$ and Gaussian noise $\boldsymbol{\epsilon}$, humanoid motion is generated through a single forward pass:
\begin{equation}
	\hat{\mathbf{x}}
	=
	f_\theta(\boldsymbol{\epsilon},\mathbf{A}).
\end{equation}

By directly learning motion-space attraction dynamics in robot configuration space, PhysDrift eliminates the generation-level embodiment gap introduced by velocity transport, produces smoother and more physically executable trajectories, and enables real-time one-step humanoid co-speech motion generation.

\section{Experiment}
\subsection{Implemented Details.} We refer to the experimental settings in Syntalker \cite{syntalker} and GestureLSM \cite{I-11}. All models used in the experiment were trained for 1000 epochs and optimized using Adam with a learning rate of 1e-4. The training process were conducted on single NVIDIA A100.

\subsection{Evaluation Metrics}

To comprehensively evaluate humanoid co-speech motion generation, we consider multiple metrics covering speech-motion synchronization, motion distribution quality, physical feasibility, motion smoothness, and real-time inference capability.

\textbf{Alignment} (Align.) \cite{II-I-2} measures the temporal alignment between speech and generated motion. Specifically, it evaluates whether motion rhythm and motion energy changes are synchronized with speech prosody and acoustic emphasis. Higher values indicate better speech-motion temporal consistency.

\textbf{Diversity} \cite{II-I-2} measures the variability of generated motions across different samples. Higher diversity indicates richer expressive motion distributions and reduced mode collapse.
	
\textbf{FMD} (Fréchet Motion Distance) evaluates the distributional similarity between generated motions and ground-truth motions. Similar to Fréchet Inception Distance (FID) \cite{fid} in image generation and Fréchet Gesture Distance (FGD) \cite{II-I-2} in gesture generation, FMD computes the Fréchet distance between feature distributions extracted from generated and real motion sequences. Lower values indicate that the generated motions are statistically closer to real humanoid motion distributions.
	
\textbf{G\_MPJPE} (Global Mean Per Joint Position Error) \cite{kungfubot} is defined as the average Euclidean distance between generated joint positions and reference joint positions in global coordinate space, measured in meters (m). Unlike local pose reconstruction metrics, G\_MPJPE additionally evaluates global spatial consistency of the reconstructed humanoid motion. Lower values indicate more accurate pose reconstruction.
	
\textbf{Joint Violation} \cite{kungfubot} is defined as the joint-angle violation beyond predefined humanoid joint limits during motion execution.

\textbf{Foot Contact Distance} \cite{omniretarget} evaluates the average height of the feet above the ground during frames that are expected to maintain foot contact, measured in meters (m). Lower values indicate more stable foot-ground contact behavior.
	
\textbf{Skating Velocity} \cite{omniretarget} quantifies the sliding velocity of the feet during contact phases, with the unit meters per second (m/s). Lower values indicate reduced foot skating artifacts and more physically plausible humanoid motion.
	
\textbf{Jerk} is defined as the third-order temporal derivative of end-effector (wrist and foot) position trajectories, corresponding to the rate of change of acceleration, measured in meters per second cubed ($m/s^3$). Lower jerk values indicate smoother and more physically stable humanoid motion dynamics. FMD, Align., and Diversity alone cannot accurately reflect high-frequency jitter; in fact, stronger jitter may even improve these metrics. Therefore, Jerk should be added as a core reference. Only when Jerk and the other three metrics are all within reasonable ranges can motion naturalness be assessed.
	
\textbf{APS} (Actions Per Second) measures the number of motion frames generated per second during inference, reflecting end-to-end latency from speech input to motion output. Higher APS indicates faster inference speed and stronger real-time interaction capability.

\subsection{Embodiment Gap in Human-Centric Co-Speech Generation}

We first analyze how retargeting affects speech-driven motion distributions in existing human-centric co-speech generation pipelines. Table \ref{tab:results} compares motion quality before and after humanoid retargeting across both ground-truth (GT) motions and generated motions from different co-speech generation models.

A consistent phenomenon can be observed across all methods: retargeting causes degradation in both speech-motion alignment and motion diversity. For example, the alignment score of GT motion decreases from 0.6897 to 0.6600 after retargeting and the diversity score drops substantially from 12.76 to 7.319. Similar trends can also be observed for Syntalker and GestureLSM.

These results suggest that modern retargeting pipelines are generally capable of preserving coarse semantic correspondence between speech and motion. However, the projection from human motion space into feasible humanoid motion space substantially compresses expressive motion distributions. Instead of causing severe kinematic failure, the embodiment gap primarily manifests as a loss of expressive variability and temporal dynamics. This observation is consistent with the formulation introduced in Section 3, where retargeting acts as a projection from the human motion manifold $\mathcal{M}_h$ into the feasible humanoid motion manifold $\mathcal{M}_r$.

Importantly, co-speech interaction is highly sensitive to such distribution compression. Human co-speech gestures rely heavily on subtle rhythmic variations, asymmetric arm dynamics, and localized motion emphasis synchronized with speech prosody. During retargeting, these expressive components are often smoothed into more conservative feasible motions, resulting in weaker conversational expressiveness despite relatively preserved semantic alignment.

\begin{table}
	\centering
	\caption{The impact of retargeting on speech motion temporal alignment and motion diversity. ``Align.*" refers to the alignment of only upper body and hand joint movements with speech. ``Syntalker" and ``GestureLSM" means data generated by Syntalker and GestureLSM model.}
	\renewcommand{\arraystretch}{1.25}
	\label{tab:results}
	\resizebox{\linewidth}{!}{
		\begin{tabular}{l|c|c|c}
			\toprule
			Data & Retargeting & Align.*$\uparrow$ & Diversity$\uparrow$ \\
			\midrule
			GT & $\times$ & 0.6897 & 12.76 \\
			GT & $\checkmark$ & 0.6600 & 7.319 \\
			\midrule
			Syntalker \cite{syntalker} & $\times$ & 0.7359 & 12.31 \\
			Syntalker & $\checkmark$ & 0.7291 & 9.527 \\
			\midrule
			GestureLSM$_{Diffusion}$ \cite{I-11} & $\times$ & 0.7384 & 12.57 \\
			GestureLSM$_{Diffusion}$ & $\checkmark$ & 0.7370 & 7.571 \\
			\midrule
			GestureLSM$_{ShortcutFlow}$ \cite{I-11} & $\times$ & 0.7490 & 12.46 \\
			GestureLSM$_{ShortcutFlow}$ & $\checkmark$ & 0.7447 & 7.525 \\
			\bottomrule
	\end{tabular}}
\end{table}

\subsection{Retargeting for Humanoid Co-Speech Motion Curation}

We further evaluate different humanoid retargeting strategies in Table \ref{tab:physics_metrics}. Unlike conventional retargeting evaluation that focuses primarily on pose reconstruction accuracy, we additionally evaluate speech-motion alignment to analyze whether retargeted motions preserve conversational dynamics required for co-speech interaction.

Existing retargeting baselines such as GMR, Mink, and PHC exhibit relatively poor physical quality and alignment performance. In particular, Mink and PHC produce severe skating artifacts with infinite skating velocity, indicating unstable contact behavior during humanoid execution. Their alignment scores also remain relatively low, ranging from 0.50 to 0.53.

Optimization-based methods substantially improve physical feasibility. Gradient-Based optimization reduces G\_MPJPE from 0.72 to 0.040. Introducing inverse-kinematics initialization and end-effector constraints further stabilizes motion quality. The slight increase in G\_MPJPE can be attributed to motion matching with high-energy audio signals, where the increased motion amplitude leads to minor violations.

However, the most important observation is that physical feasibility alone does not guarantee high-quality co-speech motion. Although Gradient-Based optimization and IK-EER w/o EE Loss achieve excellent kinematic metrics, their alignment scores remain limited at 0.55. In contrast, the full IK-EER framework improves alignment significantly to 0.69 while maintaining highly stable physical behavior. This result demonstrates that preserving speech-motion temporal synchronization is an independent and essential objective beyond conventional retargeting feasibility.

More importantly, these results highlight the role of IK-EER as a humanoid co-speech motion curation framework rather than merely a retargeting algorithm. Since robot-native co-speech generation requires large-scale physically executable humanoid motion data, preserving prosody-motion coupling during humanoid reconstruction becomes critical for dataset quality. The proposed IK-EER framework enables the construction of robot-native co-speech datasets that simultaneously preserve embodiment feasibility and conversational dynamics.

\begin{table*}[t]   
	\centering
	\caption{Retargeting Experiment. Gradient-Based refers to optimizing based solely on the remaining Physical Loss functions in Fig.\ref{fig2} while without using IK init. and EE Loss. ``NaN" means that due to the repositioning, the rear robot is in a suspended state and cannot measure the sliding speed when the foot touches the ground.Best results in \textbf{bold}, second best \underline{underlined}.}
	\small
	\renewcommand{\arraystretch}{1.2}
	\begin{tabular}{l|c|c|c|c}
			\toprule
			Method & G\_MPJPE$\downarrow$ & Foot Contact Distance$\downarrow$ & Skating Velocity$\downarrow$ & Align.$\uparrow$ \\
			\midrule
			GMR \cite{III-II-1} & 0.72 & 0.024 & 0.19 & \underline{0.59} \\
			Mink \cite{III-II-2} & 0.85 & 0.32 & NaN & 0.53 \\
			PHC \cite{PHC} & 0.88 & 0.35 & NaN & 0.50 \\
			\midrule
			Gradient-Based Optimization & \textbf{0.040} & 0.0041 & 0.072 & 0.55 \\
			IK-EER (Ours) & 0.046 & \textbf{0.0019} & \textbf{0.063} &\textbf{ 0.69} \\
			IK-EER w/o EE Loss & \underline{0.045} & \underline{0.0019} & \underline{0.063} & 0.55 \\
			\bottomrule
	\end{tabular}
	\label{tab:physics_metrics}
\end{table*}

\subsection{Effect of Motion Representation and Embodiment-Aware Generation}

Table \ref{tab:ablation} investigates the influence of motion representation and generation framework design on humanoid co-speech generation.

Using human-centric representations such as ``6D w. VQVAE" leads to relatively poor distribution quality, achieving an FMD of 4.183. Removing VQVAE slightly improves diversity from 7.254 to 9.612, but the overall motion distribution remains significantly inferior to robot-native representations. In contrast, directly modeling humanoid motion using ``6D Root $\&$ Joint Angle" dramatically improves motion distribution quality, reducing FMD to 0.6244.

These results indicate that motion representation itself strongly influences embodiment consistency. Human-centric latent representations introduce structural mismatch between learned motion distributions and feasible humanoid motion spaces, whereas robot-native joint-space representations provide a more suitable representation for humanoid co-speech behavior.

Flow-based generation further improves distribution learning capability. FM achieves the best FMD score of 0.3799 together with extremely high diversity of 29.41, demonstrating the strong expressive capability of flow matching models. However, as shown later in Table \ref{tab:com}, such unconstrained expressive generation also introduces severe jerk during humanoid execution.

Compared with purely generative flow matching, PhysDrift achieves a more balanced trade-off between motion quality, alignment, and physical stability. Although its diversity is lower than unconstrained flow generation, PhysDrift maintains substantially better embodiment consistency while preserving competitive distribution quality.

\begin{table}[t]
	\centering
	\caption{Experiment on motion representation and model architecture ablation. Best results in \textbf{bold}, second best \underline{underlined}.}
	\small
	\renewcommand{\arraystretch}{1.2}
	\resizebox{\columnwidth}{!}{%
		\begin{tabular}{l|c|c|c|c}
			\toprule
			Motion Representation & Method & FMD$\downarrow$ & Align.$\uparrow$ & Diversity$\uparrow$ \\
			\midrule
			6D w. VQVAE & Diffusion & 4.183 & \textbf{0.7593} & 7.254 \\
			6D w/o. VQVAE & Diffusion & 4.165 & \underline{0.6961} & 9.612 \\
			6D Root \& Joint Angle & Diffusion & 0.6244 & 0.6241 & 10.95 \\
			6D Root \& Joint Angle & FM & \textbf{0.3799} & 0.6746 & \textbf{29.41} \\
			6D Root \& Joint Angle & PhysDrift & \underline{0.540} & 0.6803 & \underline{11.17} \\
			\bottomrule
		\end{tabular}
	}
	\label{tab:ablation}
\end{table}

\begin{table*}[t]
	\centering
	\caption{Comparative experiment between digital human with retargeting pipeline and robot native generation. ``NFE" refers to the number of forward process steps required for inference. ``$\dagger$" indicates that the model is modified to accommodate robot motion representation. Best results in \textbf{bold}, second best \underline{underlined}.}
	\small
	\renewcommand{\arraystretch}{1.2}
	\begin{tabularx}{\textwidth}{X|c|c|c|c|c|c|c}
		\toprule
		Method & NFE & FMD$\downarrow$ & Aglin.$\uparrow$ & Diversity$\uparrow$ & Joint Violation & Jerk$\downarrow$ & APS$\uparrow$ \\
		\midrule
		GT & -- & -- & 0.5366 & 7.319 & $\times$ & 86.59 & -- \\
		\midrule
		\multicolumn{8}{c}{SMPL-X $\&$ Retargeting} \\
		\midrule
		Syntalker \cite{syntalker} $\&$ GMR \cite{III-II-1} & 1000 & 0.5633 & \underline{0.6978} & 9.527 & $\times$ & 133.9 & 11.74 \\
		GestureLSM$_{ShortcutFlow}$ \cite{I-11} $\&$ GMR \cite{III-II-1} & 2 & 0.7012 & 0.6781 & 7.769 & $\times$ & 141.6 & 40.51 \\
		GestureLSM$_{MeanFlow}$ \cite{I-11} $\&$ GMR \cite{III-II-1} & 1 & \underline{0.4241} & 0.6964 & 7.525 & $\times$ & 149.3 & 42.32 \\
		\midrule
		\multicolumn{8}{c}{Robotic Native Motion Representation} \\
		\midrule
		Syntalker$\dagger$ & 1000 & 0.6244 & 0.6241 & 10.95 & $\times$ & 141.4 & 18.11 \\
		GestureLSM$_{ShortcutFlow}\dagger$ & \underline{2} & \textbf{0.3799} & 0.6747 & \textbf{29.41} & $\checkmark$ & 975.4 & 2010 \\
		GestureLSM$_{MeanFlow}\dagger$ & 1 & 0.5542 & \textbf{0.7102} & 10.92 & $\times$ & 437.2 & \underline{2350} \\
		\midrule
		PhysDrift (Ours) & \textbf{1} & 0.5462 & 0.6856 & \underline{12.20} & $\times$ & \underline{118.0} & \textbf{2880} \\
		PhysDrift w/o. Joint Limitaion & 1 & 0.5310 & 0.6901 & 11.64 & $\checkmark$ & 118.6 & 2880 \\
		PhysDrift w/o. Joint Limitaion \& Energy Envelope & 1 & 0.5409 & 0.6803 & 11.17 & $\checkmark$ & \textbf{106.9} & 2880 \\
		\bottomrule
	\end{tabularx}
	\label{tab:com}
\end{table*}

\subsection{Overall Comparison}

Table \ref{tab:com} presents the overall comparison between human-centric pipelines, robot-native generation methods, and the proposed PhysDrift framework.

Human-centric pipelines based on SMPL-X generation followed by retargeting exhibit limited motion diversity and relatively slow inference speed. For example, ``Syntalker $\&$ GMR" achieves only 11.74 APS with a diversity score of 9.527. Although ``GestureLSM$_{MeanFlow}$ $\&$ GMR" improves generation quality and speed, the resulting motion diversity remains constrained after retargeting, consistent with the embodiment-gap analysis in Table \ref{tab:results}.

Robot-native generation substantially improves efficiency. Flow-based methods achieve extremely high inference throughput, with ``GestureLSM$_{MeanFlow}$" reaching 2350 APS. Moreover, robot-native generation also avoids the diversity collapse introduced by retargeting. However, these methods often suffer from severe physical instability. ``GestureLSM$_{ShortcutFlow}$" achieves very high diversity (29.41) and low FMD (0.3799), but simultaneously produces catastrophic joint violations and extremely high jerk (975.4), making the generated motions unsuitable for stable humanoid execution. According to the definition of Diversity, strong jitter may actually lead to higher scores. Therefore, high diversity scores alone does not necessarily indicate an advantage of the method in robot space. The high jerk of ``GestureLSM$_{ShortcutFlow}$" and ``GestureLSM$_{MeanFlow}$" also demonstrates that, without a VAE Decoder to smooth the motions, learning velocity fields is not suitable for the native motion space of robots. PhysDrift achieves a substantially better balance among expressiveness, physical plausibility, and real-time capability. Compared with unconstrained flow-based generation, PhysDrift reduces jerk from 437.2 to 118.0 while maintaining competitive diversity and alignment performance. 

The ablation variants further demonstrate the importance of embodiment-aware regularization. Removing joint constraints result in joint violations, while removing both Joint Limitation and Energy Envelope constraints reduces alignment and diversity simultaneously. These results confirm that embodiment-aware constraints are essential for stabilizing robot-native co-speech generation.

Overall, the experimental results consistently support the central hypothesis of this work: the primary limitation of existing co-speech pipelines lies not merely in physical feasibility, but in the embodiment inconsistency introduced by human-centric motion representations. By directly modeling speech-driven motion in robot joint space while incorporating embodiment-aware physical regularization, PhysDrift achieves more expressive, physically plausible, and real-time humanoid co-speech interaction.

\section{Conclusion}

In this work, we investigated the problem of humanoid co-speech motion generation from the perspective of embodiment consistency. Unlike existing human-centric pipelines that generate motion in intermediate human-body representations and subsequently retarget them onto humanoid robots, we showed that such approaches introduce a fundamental embodiment gap between human motion manifolds and physically executable humanoid motion spaces. Through extensive analysis, we demonstrated that the primary limitation of existing pipelines is not merely kinematic feasibility, but the distortion of expressive motion distributions and speech-motion temporal dynamics during embodiment transfer.

To address this problem, we proposed an embodiment-aware robot-native co-speech generation framework composed of two complementary components. First, IK-EER enables prosody-preserving humanoid motion curation by jointly optimizing kinematic feasibility and speech-motion synchronization during retargeting. Building upon the curated robot-native dataset, PhysDrift directly models speech-conditioned humanoid joint trajectories while incorporating embodiment-aware physical regularization to stabilize motion dynamics and preserve expressive behavior.

Experimental results consistently validated the proposed formulation of the embodiment gap. Our analysis showed that retargeting degraded speech-motion alignment and motion diversity, even when semantic information was largely preserved. Furthermore, while robot-native flow-based generation substantially improves expressiveness and inference efficiency, unconstrained generation often leads to unstable humanoid dynamics. By jointly modeling embodiment constraints and expressive generation within robot joint space, PhysDrift achieves a substantially better balance between motion realism, physical plausibility, speech alignment, and real-time interaction capability.

More broadly, this work suggests that humanoid co-speech generation should move beyond conventional human-centric motion representations toward embodiment-aware generative modeling directly grounded in robot morphology and dynamics. We believe that preserving embodiment consistency will become increasingly important for future socially interactive humanoid systems, where natural communication depends not only on semantic correctness, but also on physically grounded expressive motion.
\bibliographystyle{IEEEtran}
\bibliography{ref}

@inproceedings{II-I-1,
  author       = {Haiyang Liu and
                  Zihao Zhu and
                  Naoya Iwamoto and
                  Yichen Peng and
                  Zhengqing Li and
                  You Zhou and
                  Elif Bozkurt and
                  Bo Zheng},
  title        = {{BEAT:} {A} Large-Scale Semantic and Emotional Multi-modal Dataset
                  for Conversational Gestures Synthesis},
  booktitle    = {Proc. Eur. Conf. Comput. Vis., (ECCV)},
  pages        = {612--630},
  address    = {Tel Aviv, Israel},
  year         = {Oct. 2022},
}

@inproceedings{II-I-2,
  author       = {Haiyang Liu and
                  Zihao Zhu and
                  Giorgio Becherini and
                  Yichen Peng and
                  Mingyang Su and
                  You Zhou and
                  Xuefei Zhe and
                  Naoya Iwamoto and
                  Bo Zheng and
                  Michael J. Black},
  title        = {{EMAGE:} Towards Unified Holistic Co-Speech Gesture Generation via
                  Expressive Masked Audio Gesture Modeling},
  booktitle    = {Proc. IEEE Conf. Comput. Vis. Pattern Recognit., (CVPR)},
  pages        = {1144--1154},
  address    = {Seattle, WA, USA},
  year         = {Jun. 2024},
}

@article{II-I-4,
  author       = {Saeed Ghorbani and
                  Ylva Ferstl and
                  Daniel Holden and
                  Nikolaus F. Troje and
                  Marc{-}Andr{\'{e}} Carbonneau},
  title        = {ZeroEGGS: Zero-shot Example-based Gesture Generation from Speech},
  journal      = {Comput. Graph. Forum},
  volume       = {42},
  number       = {1},
  pages        = {206--216},
  year         = {2023},
}

@article{II-I-5,
  author       = {Ariel Ephrat and
                  Inbar Mosseri and
                  Oran Lang and
                  Tali Dekel and
                  Kevin Wilson and
                  Avinatan Hassidim and
                  William T. Freeman and
                  Michael Rubinstein},
  title        = {Looking to listen at the cocktail party: a speaker-independent audio-visual
                  model for speech separation},
  journal      = {{ACM} Trans. Graph.},
  volume       = {37},
  number       = {4},
  pages        = {112},
  year         = {2018},
}

@article{II-I-6,
  author       = {Youngwoo Yoon and
                  Bok Cha and
                  Joo{-}Haeng Lee and
                  Minsu Jang and
                  Jaeyeon Lee and
                  Jaehong Kim and
                  Geehyuk Lee},
  title        = {Speech gesture generation from the trimodal context of text, audio,
                  and speaker identity},
  journal      = {{ACM} Trans. Graph.},
  volume       = {39},
  number       = {6},
  pages        = {222:1--222:16},
  year         = {2020},
}

@inproceedings{II-II-1,
  author       = {Shuuji Kajita and
                  Fumio Kanehiro and
                  Kenji Kaneko and
                  Kiyoshi Fujiwara and
                  Kensuke Harada and
                  Kazuhito Yokoi and
                  Hirohisa Hirukawa},
  title        = {Biped walking pattern generation by using preview control of zero-moment
                  point},
  booktitle    = {Proc. IEEE Int. Conf. Robot. Autom., (ICRA)},
  pages        = {1620--1626},
  address    = {Taipei, Taiwan, China},
  year         = {Sep, 2003},
}

@inproceedings{II-II-2,
  author       = {Siyuan Feng and
                  Eric C. Whitman and
                  X. Xinjilefu and
                  Christopher G. Atkeson},
  title        = {Optimization based full body control for the atlas robot},
  booktitle    = {{IEEE-RAS} Int. Conf. Humanoid Rob.},
  pages        = {120--127},
  address    = {Madrid, Spain},
  year         = {Nov, 2014},
}

@article{II-II-3,
  author       = {Wael Suleiman and
                  Fumio Kanehiro and
                  Eiichi Yoshida and
                  Jean{-}Paul Laumond and
                  Andr{\'{e}} Monin},
  title        = {Time Parameterization of Humanoid-Robot Paths},
  journal      = {{IEEE} Trans. Robot.},
  volume       = {26},
  number       = {3},
  pages        = {458--468},
  year         = {2010},
}

@article{II-II-4,
  author       = {Hari Teja Kalidindi and
                  Abhilash Balachandran and
                  Suril Vijaykumar Shah},
  title        = {Optimal whole-body motion planning of humanoids in cluttered environments},
  journal      = {Robotics Auton. Syst.},
  volume       = {118},
  pages        = {263--277},
  year         = {2019},
}

@inproceedings{II-II-5,
  author       = {Marijn F. Stollenga and
                  Leo Pape and
                  Mikhail Frank and
                  J{\"{u}}rgen Leitner and
                  Alexander F{\"{o}}rster and
                  J{\"{u}}rgen Schmidhuber},
  title        = {Task-relevant roadmaps: {A} framework for humanoid motion planning},
  booktitle    = {2013 {IEEE/RSJ} Int. Conf. Intell. Robots Syst., (IROS)},
  pages        = {5772--5778},
  address    = {Tokyo, Japan},
  year         = {Nov, 2013},
}

@inproceedings{II-II-9,
  author       = {Jiageng Mao and
                  Siheng Zhao and
                  Siqi Song and
                  Chuye Hong and
                  Tianheng Shi and
                  Junjie Ye and
                  Mingtong Zhang and
                  Haoran Geng and
                  Jitendra Malik and
                  Vitor Guizilini and
                  Yue Wang},
  title        = {Universal Humanoid Robot Pose Learning from Internet Human Videos},
  booktitle    = {{IEEE-RAS} Int. Conf. Humanoid Rob.},
  pages        = {1--8},
  address    = {Seoul, Republic of Korea},
  year         = {Sep, 2025},
}

@inproceedings{II-II-6,
  author       = {Taemoon Jeong and
                  Yoonbyung Chai and
                  Sol Choi and
                  Jaewan Bak and
                  Chanwoo Kim and
                  Jihwan Yoon and
                  Yisoo Lee and
                  Jongwon Lee and
                  Kyungjae Lee and
                  Joohyung Kim and
                  Sungjoon Choi},
  title        = {CoRe: {A} Hybrid Approach of Contact-Aware Optimization and Learning
                  for Humanoid Robot Motions},
  booktitle    = {{IEEE-RAS} Int. Conf. Humanoid Rob.},
  pages        = {293--300},
  address    = {Seoul, Republic of Korea},
  year         = {Sep, 2025},
}

@inproceedings{II-II-8,
  author       = {Chenhao Lu and
                  Xuxin Cheng and
                  Jialong Li and
                  Shiqi Yang and
                  Mazeyu Ji and
                  Chengjing Yuan and
                  Ge Yang and
                  Sha Yi and
                  Xiaolong Wang},
  title        = {Mobile-TeleVision: Predictive Motion Priors for Humanoid Whole-Body
                  Control},
  booktitle    = {Proc. IEEE Int. Conf. Robot. Autom., (ICRA)},
  pages        = {5364--5371},
  address    = {Atlanta, GA, USA},
  year         = {May, 2025},
}

@inproceedings{II-II-7,
  author       = {Xuxin Cheng and
                  Yandong Ji and
                  Junming Chen and
                  Ruihan Yang and
                  Ge Yang and
                  Xiaolong Wang},
  title        = {Expressive Whole-Body Control for Humanoid Robots},
  booktitle    = {Robotics: Science and Systems XX},
  address    = {Delft, The Netherlands},
  year         = {Jul, 2024},
}

@inproceedings{I-1,
  author       = {Georgios Pavlakos and
                  Vasileios Choutas and
                  Nima Ghorbani and
                  Timo Bolkart and
                  Ahmed A. A. Osman and
                  Dimitrios Tzionas and
                  Michael J. Black},
  title        = {Expressive Body Capture: 3D Hands, Face, and Body From a Single Image},
  booktitle    = {Proc. IEEE Conf. Comput. Vis. Pattern Recognit., (CVPR)},
  pages        = {10975--10985},
  address    = {Long Beach, CA, USA},
  year         = {Jun, 2019},
}

@inproceedings{II-III-1,
  author       = {Zhenyu Jiang and
                  Yuqi Xie and
                  Jinhan Li and
                  Ye Yuan and
                  Yifeng Zhu and
                  Yuke Zhu},
  title        = {Harmon: Whole-Body Motion Generation of Humanoid Robots from Language Descriptions},
  booktitle    = {Proc. Conf. Robot Learning, (CoRL)},
  pages        = {3015--3026},
  year         = {Nov, 2024},
  address    = {Munich, Germany},
}

@article{II-III-2,
  author       = {Zihan Xu and
                  Mengxian Hu and
                  Kaiyan Xiao and
                  Qin Fang and
                  Chengju Liu and
                  Qijun Chen},
  title        = {Realizing Text-Driven Motion Generation on {NAO} Robot: {A} Reinforcement
                  Learning-Optimized Control Pipeline},
  journal      = {arXiv preprint arXiv:2506.05117},
  year         = {2025},
}

@article{II-III-3,
  author       = {Lingfan Bao and
                  Yan Pan and
                  Tianhu Peng and
                  Dimitrios Kanoulas and
                  Chengxu Zhou},
  title        = {Hierarchical Intention-Aware Expressive Motion Generation for Humanoid Robots},
  journal      = {arXiv preprint arXiv:2506.01563},
  year         = {2025},
}

@inproceedings{II-III-4,
  author       = {Yanjie Ze and
                  Zixuan Chen and
                  Wenhao Wang and
                  Tianyi Chen and
                  Xialin He and
                  Ying Yuan and
                  Xue Bin Peng and
                  Jiajun Wu},
  title        = {Generalizable Humanoid Manipulation with 3D Diffusion Policies},
  booktitle    = {Proc. {IEEE/RSJ} Int. Conf. Intell. Robots Syst.,(IROS)},
  pages        = {2873--2880},
  address    = {Hangzhou, China},
  year         = {Oct, 2025},
}

@inproceedings{II-III-5,
  author       = {Tairan He and
                  Wenli Xiao and
                  Toru Lin and
                  Zhengyi Luo and
                  Zhenjia Xu and
                  Zhenyu Jiang and
                  Jan Kautz and
                  Changliu Liu and
                  Guanya Shi and
                  Xiaolong Wang and
                  Linxi Jim Fan and
                  Yuke Zhu},
  title        = {{HOVER:} Versatile Neural Whole-Body Controller for Humanoid Robots},
  booktitle    = {Proc. IEEE Int. Conf. Robot. Autom., (ICRA)},
  pages        = {9989--9996},
  address    = {Atlanta, GA, USA},
  year         = {May, 2025},
}

@inproceedings{II-III-6,
  author       = {Zhuo Li and
                  Junjia Liu and
                  Dianxi Li and
                  Tao Teng and
                  Miao Li and
                  Sylvain Calinon and
                  Darwin G. Caldwell and
                  Fei Chen},
  title        = {ManiDP: Manipulability-Aware Diffusion Policy for Posture-Dependent
                  Bimanual Manipulation},
  booktitle    = {Proc. {IEEE/RSJ} Int. Conf. Intell. Robots Syst.,(IROS)},
  pages        = {9956--9962},
  address    = {Hangzhou, China},
  year         = {Oct, 2025},

}

@article{I-2,
author = {Widodo Budiharto and Anggita Dian Cahyani and Pingkan C.B. Rumondor and Derwin Suhartono},
title = {EduRobot: Intelligent Humanoid Robot with Natural Interaction for Education and Entertainment},
journal = {Procedia Computer Science},
volume = {116},
pages = {564-570},
year = {2017},
}

@article{I-3,
  author       = {Sara Ekstr{\"{o}}m and
                  Lena Pareto},
  title        = {The dual role of humanoid robots in education: As didactic tools and
                  social actors},
  journal      = {Educ. Inf. Technol.},
  volume       = {27},
  number       = {9},
  pages        = {12609--12644},
  year         = {2022},
}

@Article{I-4,
author="Cui, Lei
and Li, Yufei
and Yang, Xinyao
and Liu, Xue
and Zhang, Like
and Hou, Lili",
title="Humanoid Robot--Assisted Support for Health Care in Older Adults: Systematic Scoping Review",
journal="JMIR Aging",
year="2026",
volume="9",
pages="e83849",
}

@ARTICLE{I-5,
  author={Tripathi, Utkarsh and J, Rittvik Saran and Chamola, Vinay and Jolfaei, Alireza and Chintanpalli, Ananthakrishna},
  journal={IEEE Sens. J.}, 
  title={Advancing Remote Healthcare Using Humanoid and Affective Systems}, 
  year={2022},
  volume={22},
  number={18},
  pages={17606-17614},
}

@article{I-6,
  author       = {Wisanu Jutharee and
                  Boonserm Kaewkamnerdpong and
                  Thavida Maneewarn},
  title        = {Joint Reconfiguration after Failure for Performing Emblematic Gestures
                  in Humanoid Receptionist Robot},
  journal      = {Sensors},
  volume       = {23},
  number       = {22},
  pages        = {9277},
  year         = {2023},
}

@inproceedings{I-7,
  author       = {Hongye Cheng and
                  Tianyu Wang and
                  Guangsi Shi and
                  Zexing Zhao and
                  Yanwei Fu},
  title        = {{HOP:} Heterogeneous Topology-based Multimodal Entanglement for Co-Speech
                  Gesture Generation},
  booktitle    = {Proc. IEEE Conf. Comput. Vis. Pattern Recognit., (CVPR)},
  pages        = {906--916},
  address    = {Nashville, TN, USA},
  year         = {Jun, 2025}
}

@InProceedings{I-8,
    author    = {Liu, Lanmiao and Ghaleb, Esam and Ozyurek, Asli and Yumak, Zerrin},
    title     = {SemGes: Semantics-aware Co-Speech Gesture Generation using Semantic Coherence and Relevance Learning},
    booktitle = {Proc. IEEE Int. Conf. Comput. Vis., (ICCV)},
    address     = {Honolulu, Hawaii, USA},
    year      = {Oct, 2025},
    pages     = {13963-13973}
}

@inproceedings{I-9,
  author       = {Yongkang Cheng and
                  Shaoli Huang and
                  Xuelin Chen and
                  Jifeng Ning and
                  Mingming Gong},
  editor       = {Toby Walsh and
                  Julie Shah and
                  Zico Kolter},
  title        = {DIDiffGes: Decoupled Semi-Implicit Diffusion Models for Real-time
                  Gesture Generation from Speech},
  booktitle    = {Proc. AAAI Conf. Artif. Intell., (AAAI)},
  pages        = {2464--2472},
  address    = {Philadelphia, PA, USA},
  year         = {Mar, 2025},
}

@inproceedings{I-10,
  author       = {Yuxuan Bian and
                  Ailing Zeng and
                  Xuan Ju and
                  Xian Liu and
                  Zhaoyang Zhang and
                  Wei Liu and
                  Qiang Xu},
  editor       = {Toby Walsh and
                  Julie Shah and
                  Zico Kolter},
  title        = {MotionCraft: Crafting Whole-Body Motion with Plug-and-Play Multimodal
                  Controls},
  booktitle    = {Proc. AAAI Conf. Artif. Intell., (AAAI)},
  pages        = {1880--1888},
  address    = {Philadelphia, PA, USA},
  year         = {Mar, 2025},
}

@article{III-II-1,
  author       = {Jo{\~{a}}o Pedro Ara{\'{u}}jo and
                  Yanjie Ze and
                  Pei Xu and
                  Jiajun Wu and
                  C. Karen Liu},
  title        = {Retargeting Matters: General Motion Retargeting for Humanoid Motion Tracking},
  journal      = {arXiv preprint arXiv:2510.02252},
  year         = {2025},
}

@electronic{III-II-2,
    author = {Zakka, Kevin},
    title = {{Mink: Python inverse kinematics based on MuJoCo}},
    year = {2026},
    month = feb,
    version = {1.1.0},
    url = {https://github.com/kevinzakka/mink},
}

@inproceedings{I-11,
  author       = {Pinxin Liu and
                  Luchuan Song and
                  Junhua Huang and
                  Haiyang Liu and
                  Chenliang Xu},
  title        = {GestureLSM: Latent Shortcut Based Co-Speech Gesture Generation with
                  Spatial-Temporal Modeling},
  booktitle    = {Proc. IEEE Int. Conf. Comput. Vis., (ICCV)},
  pages        = {10929--10939},
  address = {Honolulu, HI, USA},
  publisher    = {{IEEE}},
  year         = {Oct, 2025},
}

@inproceedings{syntalker,
  author = {Bohong Chen and Yumeng Li and Yao-Xiang Ding and Tianjia Shao and Kun Zhou},
  title = {Enabling Synergistic Full-Body Control in Prompt-Based Co-Speech Motion Generation},
  booktitle = {Proceedings of the 32nd ACM International Conference on Multimedia},
  year = {2024},
  publisher = {ACM},
  address = {New York, NY, USA},
  pages = {10},
  doi = {10.1145/3664647.3680847}
}

@article{I-12,
  author       = {Haozhe Jia and
                  Jianfei Song and
                  Yuan Zhang and
                  Honglei Jin and
                  Youcheng Fan and
                  Wenshuo Chen and
                  Wei Zhang and
                  Yutao Yue},
  title        = {{ECHO:} Edge-Cloud Humanoid Orchestration for Language-to-Motion Control},
  journal      = {arXiv preprint arXiv:2603.16188},
  year         = {2026},
}

@article{phuma,
  author       = {Kyungmin Lee and
                  Sibeen Kim and
                  Minho Park and
                  Hyunseung Kim and
                  Dongyoon Hwang and
                  Hojoon Lee and
                  Jaegul Choo},
  title        = {{PHUMA:} Physically-Grounded Humanoid Locomotion Dataset},
  journal      = {arXiv preprint arXiv:2510.26236},
  year         = {2025},

}

@article{ncc,
title = {Adaptive convolutional network pruning through pixel-level cross-correlation and channel independence for enhanced model compression},
journal = { Eng. Appl. Artif. Intell.},
volume = {154},
pages = {110920},
year = {2025},
issn = {0952-1976},
author = {Guangyao Li and Haijian Shao and Xing Deng and Yingtao Jiang},}

@article{drift,
  author       = {Mingyang Deng and
                  He Li and
                  Tianhong Li and
                  Yilun Du and
                  Kaiming He},
  title        = {Generative Modeling via Drifting},
  journal      = {arXiv preprint arXiv:2602.04770},
  year         = {2026},
}

@inproceedings{PHC,
  author       = {Zhengyi Luo and
                  Jinkun Cao and
                  Alexander Winkler and
                  Kris Kitani and
                  Weipeng Xu},
  title        = {Perpetual Humanoid Control for Real-time Simulated Avatars},
  booktitle    = {Proc. IEEE Int. Conf. Comput. Vis., (ICCV)},
  pages        = {10861--10870},
  address = {Paris, France,},
  year         = {Oct, 2023},
}

@inproceedings{6d,
  author       = {Yi Zhou and
                  Connelly Barnes and
                  Jingwan Lu and
                  Jimei Yang and
                  Hao Li},
  title        = {On the Continuity of Rotation Representations in Neural Networks},
  booktitle    = {Proc. IEEE Conf. Comput. Vis. Pattern Recognit., (CVPR)},
  pages        = {5745--5753},
  address    = {Long Beach, CA, USA},
  year         = {Jun, 2019}
}

@inproceedings{fm,
  author       = {Yaron Lipman and
                  Ricky T. Q. Chen and
                  Heli Ben{-}Hamu and
                  Maximilian Nickel and
                  Matthew Le},
  title        = {Flow Matching for Generative Modeling},
  booktitle    = {Proc. Int. Conf. Learn. Represent.,(ICLR)},
  address    = {Kigali, Rwanda},
  year         = {May, 2023},
}

@inproceedings{fid,
author = {Heusel, Martin and Ramsauer, Hubert and Unterthiner, Thomas and Nessler, Bernhard and Hochreiter, Sepp},
title = {GANs trained by a two time-scale update rule converge to a local nash equilibrium},
year = {Dec, 2017},
address = {Long Beach, California, USA},
booktitle = {Proc. Adv. neural inf. proces. syst., (NeurIPS)},
pages = {6629-6640},
numpages = {12},
}

@inproceedings{kungfubot,
  title={KungfuBot: Physics-Based Humanoid Whole-Body Control for Learning Highly-Dynamic Skills},
  author={Xie, Weiji and Han, Jinrui and Zheng, Jiakun and Li, Huanyu and Liu, Xinzhe and Shi, Jiyuan and Zhang, Weinan and Bai, Chenjia and Li, Xuelong},
  booktitle = {Proc. Adv. neural inf. proces. syst., (NeurIPS)},
  pages={62406--62433},
  address = {San Diego, CA, USA},
  year={2025}
}

@article{omniretarget,
  title={Omniretarget: Interaction-preserving data generation for humanoid whole-body loco-manipulation and scene interaction},
  author={Yang, Lujie and Huang, Xiaoyu and Wu, Zhen and Kanazawa, Angjoo and Abbeel, Pieter and Sferrazza, Carmelo and Liu, C Karen and Duan, Rocky and Shi, Guanya},
  journal={arXiv preprint arXiv:2509.26633},
  year={2025}
}

@inproceedings{vqvae,
author = {van den Oord, Aaron and Vinyals, Oriol and Kavukcuoglu, Koray},
title = {Neural discrete representation learning},
year = {2017},
address = {Long Beach, CA, USA},
booktitle = {Proc. Adv. neural inf. proces. syst., (NeurIPS)},
pages = {6309-6318},
}

@inproceedings{livegesture,
  title={LiveGesture: Streamable Co-Speech Gesture Generation Model},
  author={Saleem, Muhammad Usama and Patel, Mayur Jagdishbhai and Pinyoanuntapong, Ekkasit and Qin, Zhongxing and Yang, Li and Xue, Hongfei and Helmy, Ahmed and Chen, Chen and Wang, Pu},
  booktitle = {Proc. IEEE Conf. Comput. Vis. Pattern Recognit., (CVPR)},
  pages={2264--2273},
  address = {Denver, Colorado, USA},
  year={Jun, 2026}
}

@InProceedings{CoordSpeaker,
    author    = {Fang, Fengyi and Yang, Sicheng and Yang, Wenming},
    title     = {CoordSpeaker: Exploiting Gesture Captioning for Coordinated Caption-Empowered Co-Speech Gesture Generation},
    booktitle = {Proc. IEEE Conf. Comput. Vis. Pattern Recognit., (CVPR)},
    address = {Denver, Colorado, USA},
    year={Jun, 2026},
    pages     = {30761-30771}
}

@ARTICLE{UniTracker,
  author={Yin, Kangning and Zeng, Weishuai and Fan, Ke and Dai, Minyue and Wang, Zirui and Zhang, Qiang and Tian, Zheng and Wang, Jingbo and Pang, Jiangmiao and Zhang, Weinan},
  journal={IEEE Rob. Autom. Lett. }, 
  title={UniTracker: Learning Universal Whole-Body Motion Tracker for Humanoid Robots}, 
  year={2026},
  volume={11},
  number={7},
  pages={8124-8131},
  }

@inproceedings{humanmimic,
  title={Humanmimic: Learning natural locomotion and transitions for humanoid robot via wasserstein adversarial imitation},
  author={Tang, Annan and Hiraoka, Takuma and Hiraoka, Naoki and Shi, Fan and Kawaharazuka, Kento and Kojima, Kunio and Okada, Kei and Inaba, Masayuki},
  pages={13107--13114},
  year={May, 2024},
  booktitle    = {Proc. IEEE Int. Conf. Robot. Autom., (ICRA)},
  address    = {Yokohama, Japan},
}

@ARTICLE{EmotionGesture,
  author={Qi, Xingqun and Liu, Chen and Li, Lincheng and Hou, Jie and Xin, Haoran and Yu, Xin},
  journal={IEEE Trans. Multimedia}, 
  title={EmotionGesture: Audio-Driven Diverse Emotional Co-Speech 3D Gesture Generation}, 
  year={2024},
  volume={26},
  number={},
  pages={10420-10430},
  }

@ARTICLE{MambaGesture2,
  author={Fu, Chencan and Wang, Yabiao and He, Haoyang and Wang, Shuo and Wang, Chengjie and Tai, Ying and Liu, Yong and Zhang, Jiangning},
  journal={IEEE Trans. Multimedia}, 
  title={MambaGesture2: Co-Speech Gesture Generation via Hierarchical Fusion and Spatiotemporal Aggregation}, 
  year={2026},
  volume={Early Access},
  pages={1-9},}

\newpage

\vfill

\end{document}